\documentclass[10pt,twocolumn,letterpaper]{article}

\usepackage{cvpr}              % To produce the CAMERA-READY version

\usepackage[accsupp]{axessibility}  % Improves PDF readability for those with disabilities.
\usepackage{xcolor}

\definecolor{cvprblue}{rgb}{0.21,0.49,0.74}
\usepackage[pagebackref,breaklinks,colorlinks,allcolors=cvprblue]{hyperref}

\def\paperID{1001} % *** Enter the Paper ID here
\def\confName{CVPR}
\def\confYear{2025}

\title{D$^2$iT: Dynamic Diffusion Transformer for Accurate Image Generation}

\author{Weinan Jia\textsuperscript{\rm 1}, Mengqi Huang\textsuperscript{\rm 1},  Nan Chen\textsuperscript{\rm 1}, Lei Zhang\textsuperscript{\rm 1}, Zhendong Mao\textsuperscript{\rm 1 2}\thanks{Zhendong Mao is the corresponding author.} \\
\textsuperscript{\rm 1}University of Science and Technology of China,
Hefei, China; \\
\textsuperscript{\rm 2}Institute of Artificial intelligence, Hefei Comprehensive National Science Center, Hefei, China \\
{\tt\small \{jiawn, huangmq, chen\_nan\}@mail.ustc.edu.cn, \{leizh23, zdmao\}@ustc.edu.cn}}

\begin{document}
\maketitle
\begin{abstract}
Diffusion models are widely recognized for their ability to generate high-fidelity images. Despite the excellent performance and scalability of the Diffusion Transformer (DiT) architecture, it applies fixed compression across different image regions during the diffusion process, disregarding the naturally varying information densities present in these regions. However, large compression leads to limited local realism, while small compression increases computational complexity and compromises global consistency, ultimately impacting the quality of generated images. To address these limitations, we propose dynamically compressing different image regions by recognizing the importance of different regions, and introduce a novel two-stage framework designed to enhance the effectiveness and efficiency of image generation: (1) Dynamic VAE (DVAE) at first stage employs a hierarchical encoder to encode different image regions at different downsampling rates, tailored to their specific information densities, thereby providing more accurate and natural latent codes for the diffusion process. (2) Dynamic Diffusion Transformer (D$^2$iT) at second stage generates images by predicting multi-grained noise, consisting of coarse-grained (less latent code in smooth regions) and fine-grained (more latent codes in detailed regions), through an novel combination of the Dynamic Grain Transformer and the Dynamic Content Transformer. The strategy of combining rough prediction of noise with detailed regions correction achieves a unification of global consistency and local realism. Comprehensive experiments on various generation tasks validate the effectiveness of our approach. Code will be released at \url{https://github.com/jiawn-creator/Dynamic-DiT}.
\end{abstract}

% We conduct comprehensive experiments on the ImageNet 256$\times$256 benchmark, showing that D$^2$iT achieves 23.8\% quality improvement than DiT (D$^2$iT's 1.73 \vs DiT's 2.27 on FID score, lower better), by using only 57.1\% of the computational resources as DiT.    
\section{Introduction}
\label{sec:intro}

%Recent advancements in deep learning for computer vision have significantly improved the quality of image generation. Diffusion models, in particular, have become essential tools capable of transforming various inputs (\eg, class labels, text descriptions, and sketches) into highly realistic images. Despite these developments, a central research goal remains: ensuring the authenticity of local details while maintaining the consistency of the global structure in generated images.
\begin{figure}[t]
  \centering
   \includegraphics[width=1\linewidth]{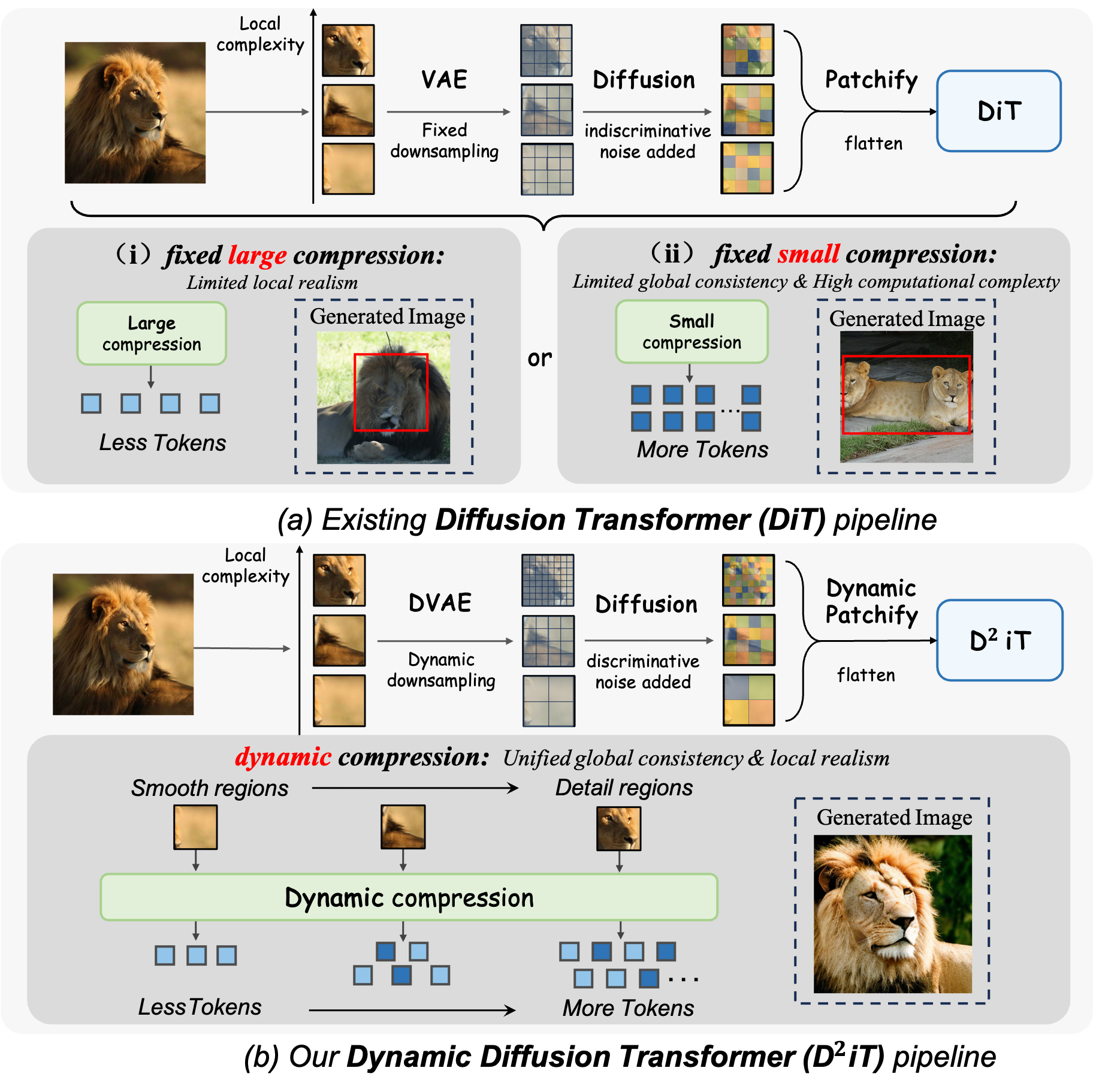}

   \caption{Illustration of our motivation. Compression here refers to the VAE + Patchify operation. (a) Existing fixed-compression diffusion transformer (DiT) \textbf{ignore information density}. Fixed large compression leads to limited local realism due to the limited representation of a few tokens preventing accurate recovery of rich information, whereas fixed small compression leads to limited global consistency and high computational complexity due to the burden of global modeling across patched latents. Samples in (a) are obtained from  \cite{peebles2023scalable}. (b) Our Dynamic Diffusion Transformer (D$^2$iT) adopts a \textbf{dynamic compression strategy} and adds multi-grained noise based on information density, achieving unified global consistency and local realism.  } 
   \label{fig:motivation}
\end{figure}

In recent years, vision generative models have advanced significantly, raising the realism and fidelity of visual generation to new heights.
Among them, the Diffusion Transformer (DiT) has attracted considerable attention and become the de facto choice for many modern image and video generative models such as Stable Diffusion 3  \cite{esser2024scaling}, Flux  \cite{Black_Forest_Labs} and CogvideoX  \cite{yang2024cogvideox}, primarily because it combines the best of both worlds, \ie, the scalability of transformer architecture and the powerful modeling of diffusion process.

Due to the high computational cost in pixel space and challenges of capturing  high-level semantic features, existing DiT-based generation models \cite{peebles2023scalable,Gao_2023_ICCV,zheng2023fast,chen2023pixart,li2024hunyuan} generally follow a two-stage paradigm, \ie, (1) in the first stage, a Variational Autoencoder (VAE)  \cite{kingma2013auto} is utilized to spatially compress the image into a low-dimensional latent space representation; 
(2) in the second stage, this latent representation is further spatially patched to form a more compressed one,
which is then modeled by the diffusion process within a transformer architecture. 
\emph{The key to the success of DiT lies in its spatial compression}, which significantly reduces the image sequence length. This reduction is essential for the transformer's self-attention mechanism to model global structures, while not markedly compromising local details. 

Recent improvements to DiT primarily focus on accelerating convergence and broadening its applicability to various downstream tasks.  Approaches such as MDT \cite{Gao_2023_ICCV} and Mask DiT \cite{zheng2023fast} employ masking strategies to speed up training convergence and enhance the model's capacity for representative learning. FiT \cite{lu2024fit} utilizes rotational position encoding to enable variable-resolution capabilities, allowing the model to generate images of different sizes. 
Furthermore, industrial-scale models such as Pixart-$\alpha$ \cite{chen2023pixart} and Hunyuan-DiT \cite{li2024hunyuan} incorporate cross-attention mechanisms and advanced text encoders like CLIP \cite{radford2021learning} and T5 \cite{raffel2020exploring}, achieving high-quality, text-guided image generation using large datasets and effective training strategies. 
% Existing DiT-based methods treat all image regions equally during diffusion process and patchify operation, without considering the differences in information density in different regions.
Though great progress has been made, the commonality among existing DiT-based methods is that they all \emph{leave the key compression principle untouched}, \ie, using a fixed downsampling ratio for all image regions equally. Specifically, a fixed  pixel region is compressed into a latent token, which is then diffused and denoised for visual generative modeling without considering the information density of different regions.

In this study, we argue that the fixed compression employed by existing DiT-based models \emph{overlooks the natural variation in spatial complexity} across different image regions. 
As a result, these models are constrained in integrating consistent global visual structures with realistic local visual details, and suffer from slow training convergence.
The root cause of this limitation is that the image diffusion transformer model, by its very nature, learns to progressively recover each region's image information from a pure Gaussian noisy patched latent through the built-in self-attention mechanism. 
On the one hand, while a large fixed compression with a short image sequence is effective for the self-attention mechanism to capture dependencies across patched latents, it fails to accurately recover all the rich information for detailed regions due to their high spatial complexity. As shown in Figure \ref{fig:motivation} (a) (i), the large compressed tokens are overwhelmed to model the details and do not guarantee the realism of the lion's face.  %\hmq{[need detailed explanation based on the image]}. 
On the other hand, the small fixed compression results in a much longer image sequence, which can better recover the local details within each region but significantly increases the computational difficulty and burden for consistent global modeling across patched latents. As shown in Figure \ref{fig:motivation} (a) (ii), despite the realistic lion's face in the generated image, its body structure has obvious defects due to inaccurate global modeling of the long image sequence.  %\hmq{[need detailed explanation based on the image]}.

To address the above challenges, we propose \emph{Dynamic Diffusion Transformer} (D$^2$iT) for accurate image representation modeling both across and within patched latents, achieving the integration of consistent global structures and realistic local details with faster training convergence and a smaller computational burden.
As shown in Figure \ref{fig:motivation} (b), the key idea of D$^2$iT is to adaptively compress image regions into different grained representations with various numbers of latents.
Specifically, we introduce a novel two-stage framework, \ie, (1) \emph{Dynamic VAE} (DVAE) is used in the first stage to encode image regions into different downsampling continuous latent representations according to their different spatial complexity;
% which is achieved by a hierarchical encoder and a \emph{Dynamic Granularity Coding (DGC)} module;
% The hierarchical encoder provides multiple latent code candidates using a series of downsampling rates, and the DGC module assigns the most appropriate latent representation based on the information density of different region; 
% (2) D$^2$iT includes Dynamic Grain Transformer for modeling spatial information and Dynamic Content Transformer for modeling noise content.
% The Dynamic Grain Transformer learns the real space information density distribution from the DVAE and predicts grain map to guide Dynamic Content Transformer as a condition.
% Subsequently, Dynamic Content Transformer applies varying levels of noise to different regions based on their assigned granularity (small compression noise for detailed regions and large compression noise for smooth regions).
% Specifically, a \emph{Multi-grained Net} (large patch size) + \emph{Fine-grained Net} (small patch size) is designed to maintain global consistency and improve detail realism in Dynamic Content Transformer. 
% Multi-grained Net ensures global consistency and predicts mixture noise roughly. 
% The Fine-grained Net performs noise correction for detail-rich regions that are roughly predicted by the Multi-grained Net.
(2) D$^2$iT is proposed to complete a multi-grained noise prediction task. Considering that the natural image regions inherently contain spatial density information and image content information, multi-grained noise prediction can be decomposed into two parts: spatial information density modeling and content information modeling. Therefore, we designed a \textit{Dynamic Grain Transformer} for modeling spatial information density and a \textit{Dynamic  Content Transformer} for modeling noise content. The Dynamic Grain Transformer learns the true spatial density distribution from DVAE and predicts a grain map. Then, the Dynamic Content Transformer applies different compression levels of noise to different regions according to their assigned granularities. In order to maintain global consistency and local details, we adopt a global noise rough prediction and fine-grained noise correction strategy in Dynamic Content Transformer. 
% The efficient fine-grained Net is designed to correct noise in detailed regions, as roughly predicted by previous.

In summary, our main contributions are as follows:

\textcolor{blue}{\textbf{Conceptual Contributions}}. We identified that existing diffusion processes \textbf{overlook the natural variation} in spatial complexity across different image regions, leading to limitations in integrating a consistent global visual structure and realistic local visual details. We propose a more naturally \textbf{information-density-based} diffusion process and dynamic information compression.

\textcolor{blue}{\textbf{Technical Contributions}.} We propose a novel two-stage generation architecture. The proposed D$^2$iT generates image by predicting multi-grained noise. The novel rough prediction and fine-grained correction strategy make the diffusion process more efficient. 

% We propose a novel two-stage generation architecture. In the first stage, DVAE is used to dynamically compress and encode regions according to information density. In the second stage, D$^2$iT is proposed to generate images. The novel coarse prediction and fine-grained correction strategy makes the diffusion process more efficient.

\textcolor{blue}{\textbf{Experimental Contributions}.} 
We used a similar number of parameters and only 57.1\% of the training resources to achieve 23.8\% quality improvement compared to DiT on the ImageNet Dataset, \ie, our 1.73 FID \vs DiT's 2.27 FID. 

% Our experimental results ache FID1.90 on ImageNet dataset using only 36\% of the computational consumption compared to DiT

\section{Related Works}
\label{sec:formatting}

\begin{figure*}
  \centering
   \includegraphics[width=1\linewidth]{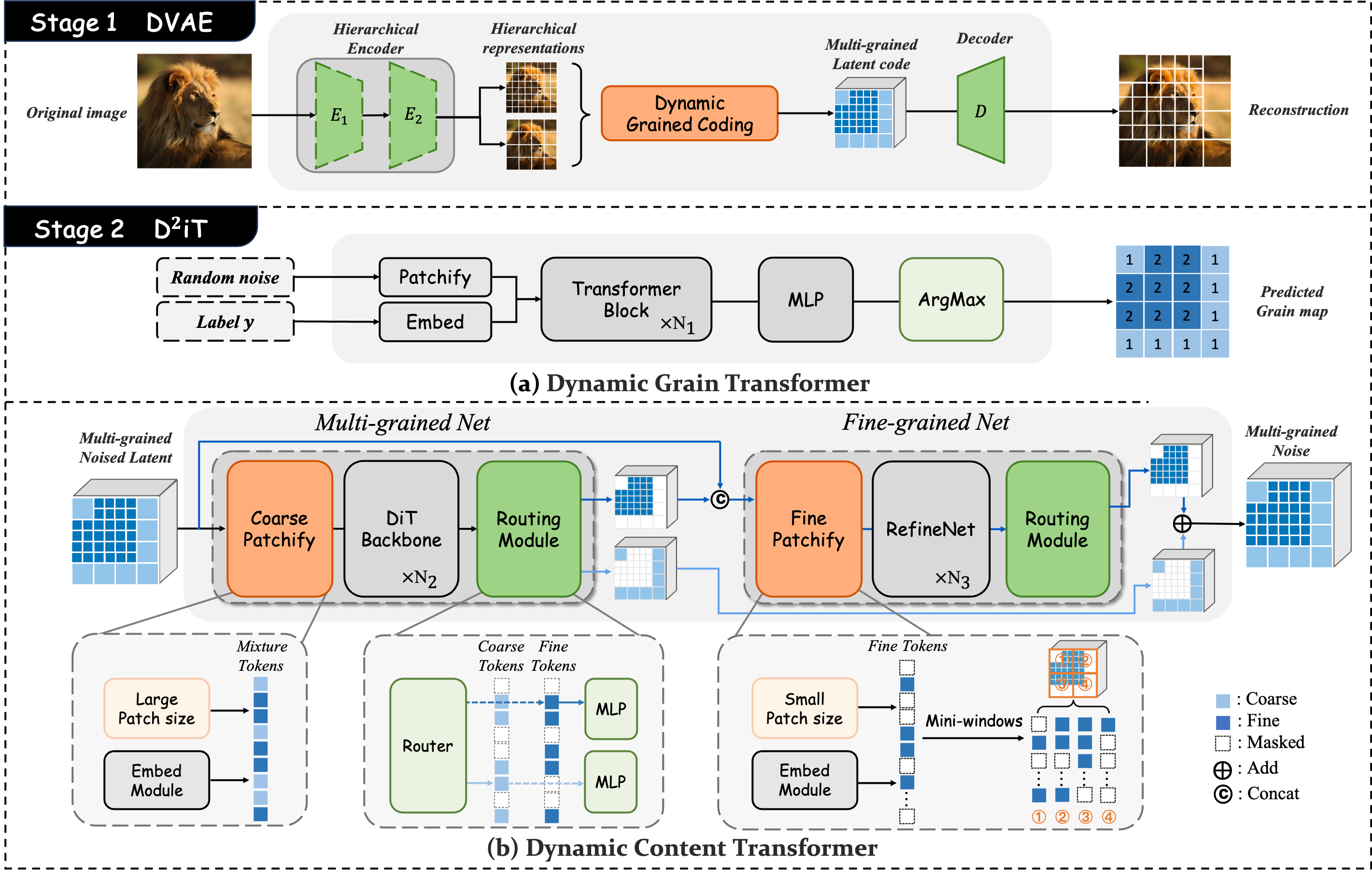}
  \caption{The overview of our proposed two-stage framework.
(1) Stage 1: DVAE dynamically assigns different grained codes to each image region through the Herarchical Encoder and Dynamic Grained Coding (DGC) module.
(2) Stage 2: D$^2$iT consists Dynamic Grain Transformer and Dynamic Content Transformer, which respectively model the spatial granularity information and content information. We present the network with two granularities. The grain map uses `1' to denote coarse-grained regions and `2' for fine-grained regions.   }
%Routing Module divides the output of the stacked DiT backbone into coarse grain tokens and fine grain tokens, and passes them through different MLPs to generate prediction content of coarseness and fineness. The fine grain part is further updated through the Fine-grained Efficient RefineNet.
  \label{fig:framework}
\end{figure*}
%-------------------------------------------------------------------------
\subsection{Variational Autoencoder for Generation Model}

Variational Autoencoder (VAE) \cite{esser2021taming} is a compression coding model that can represent information such as images and videos more compactly in a latent space. Most current mainstream image and video generation models adopt a two-stage paradigm. In the first stage, a VAE compresses and encodes the image or video into the latent space. In the second stage, the image or video distribution is remodeled within this low-dimensional latent space. Thanks to the compact information representation, this two-stage generation approach has become prevalent and is utilized by many milestone models, such as DALL-E \cite{ramesh2021zero}, latent diffusion \cite{rombach2022high}, Sora \cite{openaisora}, \textit{etc}. However, this fixed-length coding does not consider information density. Regions rich in detail have the same number of coding representations as background areas, leading to insufficient and redundant coding. To address this problem, \cite{huang2023towards} proposed a dynamic coding VAE and achieved success in the autoregressive model. Diffusion models using fixed-length coding also overlook information density. Our work refines this approach and is the first to transfer the concept of dynamic coding to diffusion models in a continuous space.

%-------------------------------------------------------------------------

\subsection{Diffusion Model}
Diffusion models \cite{sohl2015deep,ho2020denoising,song2019generative} are effective generative models. The noise prediction network gradually generates denoised samples of the input data by learning the denoising process. They have demonstrated significant potential in generating high-quality images  \cite{nichol2021glide,ramesh2022hierarchical,rombach2022high,saharia2022photorealistic, karras2024guiding} and have advanced progress in various fields  \cite{li2022diffusion,kong2020diffwave,vahdat2022lion,xu2022geodiff, chen2024customcontrast, chen2024dreamidentity, huang2024realcustom1, mao2024realcustom}. Recent improvements in sampling methods \cite{song2020denoising,karras2022elucidating,lu2022dpm} and the classifier-free guidance \cite{ho2022classifier} have further enhanced their capabilities. Latent Diffusion Models (LDMs) \cite{rombach2022high,peebles2023scalable,podell2023sdxl} adopts a two-stage generative architecture that efficiently encodes images using pre-trained autoencoders  \cite{kingma2013auto} and performs the diffusion process in a low-dimensional latent space. This approach addresses the challenges associated with generating images directly in pixel space. In this study, we develop a dynamic grained denoising network and verify its applicability in DiT framework.

\section{Method}
The Diffusion Transformer (DiT) \cite{peebles2023scalable}  employs a two-stage generation framework that compresses fixed pixel regions into tokens, allowing the transformer framework to model diffusion noise. However, the uniform treatment of all regions makes it difficult to distinguish between detailed regions and smooth background, hindering accurate modeling of global consistency of entire image and details in high-information regions. This leads to suboptimal results.

Considering that the natural images possess varying perceptually important regions and diverse spatial distributions, 
we design a two-stage framework shown in Figure \ref{fig:framework} to learn the dynamic priors of the natural images. The first stage DVAE (Section \ref{DVAE}) aims to encode the image more accurately  by identifying the information density and using different downsampling rates for different regions, and the second stage D$^2$iT (Section \ref{DiT}) learns the spatial density information and content information of the multi-grained codes to generate the image more naturally.
% An overview of our two-stage framework is illustrated in Figure \ref{fig:framework}. We provide a brief preliminary to the adopted baseline, a diffusion model based on the Transformer architecture, followed by a detailed description of our method, which comprises two components: DVAE (Section \ref{DVAE}) and D$^2$iT (Section \ref{DiT}). 

% The Diffusion Transformer (DiT) \cite{peebles2023scalable} leverages the Transformer architecture to replace the U-Net architecture commonly used in prior diffusion models. By inheriting the successes of the Transformer in various tasks, DiT demonstrates superior generative performance compared to U-Net at the same scale. DiT employs a two-stage generation framework. In the first stage, Variational Autoencoders (VAE) encode images into latent representations with fixed downsampling. In the second stage, the patchify operation further downscales the images. However, this uniform treatment to all regions makes it difficult to distinguish between high detail regions and less informative background. In the process of image generation, this uniform treatment disperses the model's focus across many low-information regions, and it is difficult to accurately model the details of high-information regions, leading to suboptimal results. 

\subsection{Dynamic VAE (DVAE)}
\label{DVAE}
Different from existing VAE-based works \cite{esser2021taming, lee2022autoregressive, yu2021vector} that use a fixed downsampling factor $f$ to represent image regions with a fixed encoding length, our \textit{Dynamic VAE} (DVAE) first defines a set of $k$ staged downsampling factors \{$f_1,f_2,\cdots,f_k$\}, where $f_1 < f_2 < \cdots < f_k$. 
As shown in Stage 1 of Figure \ref{fig:framework}, an image $\boldsymbol{X}\in \mathbb{R}^{H_0 \times W_0 \times 3}$ is first encoded into grid features $\boldsymbol{Z} = \{\boldsymbol{Z}_1, \boldsymbol{Z}_2, \cdots, \boldsymbol{Z}_k\}$ through the hierarchical encoder $E_h$, where $\boldsymbol{Z}_i \in \mathbb{R}^{H_i \times W_i \times n_z}$ and the shape $(H_i,W_i)$ is defined as:
\begin{equation}
  (H_i,W_i) = (H_0 / f_i, W_0 / f_i), i \in \{1,2,\cdots,k\}.
  \label{eq:Dynamic-VAE1}
\end{equation}

Using the maximum downsampling factor $f_k$,  the original image is segmented into regions with the size $S^2 = f_k^2$, \ie, a total of $N_p = H_0/S \times W_0/S$ regions. 

Subsequently, the \textit{Dynamic Grained Coding} module allocates the most suitable granularity to each region based on the local information entropy, resulting in a multi-grained latent representation. The Dynamic Grained Coding module employs Gaussian kernel density estimation to analyze pixel intensity distributions within each region and uses Shannon entropy to quantify the complexity of each region.
To handle the irregular latent code with different grained regions, we further propose a simple and effective neighbor copying method. Specifically, the latent code for each region is copied to the finest granularity of codes if the finest granularity is not assigned for it. 

\textbf{Dynamic Grained Coding.} Inspired by the discrete version of DQVAE\footnote{Code of discrete version is released at \url{https://github.com/CrossmodalGroup/DynamicVectorQuantization}.}\cite{huang2023towards}, the Dynamic Grained Coding module begins by converting the original image into a single-channel image denoted as $\boldsymbol{Y}\in \mathbb{R}^{H_0 \times W_0 \times 1}$. Then, the single-channel image $Y$ is divided into non-overlapping regions, each of size $S\times S$.  To assess the local information content of each region, the Dynamic Grained Coding module employs Gaussian kernel density estimation to compute the probability density function (PDF) $\hat{p}_k(\cdot)$ of pixel intensities within the $k$-th region:
\begin{equation} 
\hat{p}_k(b_j) = \frac{1}{S^2} \sum_{i=1}^{S^2} \exp\left( -\frac{1}{2} \left( \frac{x_{k,i} - b_j}{\sigma} \right)^2 \right) ,
\label{eq_pdf} 
\end{equation}
where $x_{k,i}$ denotes the $i$-th pixel value in the $k$-th region, $\sigma = 0.01$ is the smoothing parameter, and $\{b_j\}_{j=1}^P s$ represents a set of histograms uniformly distributed, where $P = S \times S$  denotes the total number of pixels in each region. Subsequently, the entropy $E_k$ of each region is calculated using Shannon's entropy formula \cite{lin1991divergence}:
\begin{equation} E_k = -\sum_{j=1}^M \hat{p}_k(b_j) \log \hat{p}_k(b_j). \label{eq_Shannon} 
\end{equation}

These entropy values are then assembled into an entropy map $\boldsymbol{E} \in \mathbb{R}^{H_0/S \times W_0/S}$. To determine the appropriate processing granularity for each region, we pre-calculate the entropy distribution of natural images in the ImageNet dataset \cite{deng2009imagenet}. This allows us to establish entropy thresholds corresponding to specific percentiles of information content. By specifying a set of desired grained ratios $r = \{r_1, r_2, \cdots, r_k\}$, we select corresponding entropy thresholds $T = \{T_1, T_2, \cdots, T_k\}$ such that the proportion of image regions with entropy values exceeding $T_i$ matches the grained ratio $r_i$. This ensures that a ratio $r_i$ of the regions are assigned to granularity $f_i$, where regions with higher entropy undergo finer-grained processing, and those with lower entropy receive coarser-grained treatment.

% \begin{figure}[t]
%   \centering
%    \includegraphics[width=1\linewidth]{images/stage1.png}

%    \caption{Example of caption.
%    It is set in Roman so that mathematics (always set in Roman: $B \sin A = A \sin B$) may be included without an ugly clash.}
%    \label{fig:stage1}
% \end{figure}

\subsection{Dynamic Diffusion Transformer (D$^2$iT)}
\label{DiT}
\subsubsection{Overview of Multi-grained Diffusion.}

% Natural images possess varying perceptually important regions and diverse spatial distributions, making the learning of dynamic latent priors highly challenging. This complexity offers a promising research direction for diffusion models, specifically in achieving natural and more efficient multi-grained image generation. Leveraging DVAE, which inherently separates coarse regions (smooth regions requiring fewer codes) from fine regions (detail-rich regions requiring more codes). 
% The accurate encoding of the DVAE has both spatial grain information and content information for each image.

The multi-grained diffusion process can be conceptualized as two consecutive steps. First, we predict the spatial distribution of information complexity (grain map) across the image using the \emph{Dynamic Grain Transformer}. Then, we perform the multi-grained diffusion process within this naturally informed spatial distribution using the \emph{Dynamic Content Transformer}, allowing for a better representation of the inherent characteristics of natural images.

\subsubsection{Dynamic Grain Transformer}

% To implement a dynamic granularity diffusion model, we need to model the spatial distribution of image granularity. 
The first goal of D$^2$iT is to model the spatial granularity distribution and predict the grain map $\boldsymbol{M}\in \mathbb{R}^{(H_0/S) \times (W_0/S)}$ for the entire image.  To achieve this, as shown in Stage 2 (a) of Figure \ref{fig:framework}, we employ a \textit{Dynamic Grain Transformer}, which generates the grain map $\boldsymbol{M}$ by sampling from random noise, thereby effectively capturing the global granularity distribution throughout the image. The ground truth grain map used for training originates from the outputs of the Dynamic Grained Coding  module within the DVAE. Specifically, for the patch region in row $i$ and column $j$, the granularity $\theta_{i,j}$ is determined by the downsampling factor utilized during the reconstruction phase of the DVAE, reflecting the appropriate granularity for that specific region:
\begin{equation} \theta_{i,j} = \arg\max_{l} (\boldsymbol{g}_{i,j,l})\in \{ 1,2,...,k\},
\label{eq:position}
\end{equation}
where  $\boldsymbol{g}_{i,j,l}$  is the predicted probability of granularity $l$ for each region, $1 < i < H_0/S, 1 < j <W_0/S$ , and $l\in\{1, 2, \dots, k\}$.
The cross-entropy loss function for Dynamic Grain Transformer training is:
\begin{equation} \mathcal{L}_{\text{grain}} = - \mathbb{E}_{i,j} \sum_{l=1}^K y_{i,j,l} \log ( \boldsymbol{g}_{i,j,l} ), \label{eq
} \end{equation}
where  $y_{i,j,l}$ is the ground truth granularity distribution at region $(i,j)$. %, represented as a one-hot vector:
% \begin{equation} 
% y_{i,j,l} = 
% \begin{cases} 
% 1, & \text{if } \theta_{i,j} = l \\
% 0, & \text{otherwise} \\
% \end{cases}.
% \label{eq_g_ijl
% } \end{equation}
By learning a real spatial grain map, the Dynamic Grain Transformer can effectively model the spatial distribution of information complexity. This grain map $\boldsymbol{M}$ is then utilized to guide the multi-grained diffusion process.

\subsubsection{Dynamic Content Transformer}

After obtaining the spatial information of the image, the next step is to model the content information. We propose the \textit{Dynamic Content Transformer} for the task of multi-grained noise prediction. 
In order to align with the patchify operation of existing SOTA methods \cite{peebles2023scalable},which generally adopt a patch size of 2, we present a dual-grained network (coarse with patch size of 2 and fine with patch size of 1). As depicted in Stage 2 (b) of Figure \ref{fig:framework}, the Dynamic Content Transformer consists of a \emph{Multi-grained Net} and a \emph{Fine-grained Net}, which realizes multi-grained noise rough prediction and fine-grained noise correction. First, multi-grained Net tokenizes $\boldsymbol{z}_{\text{noised}}$ using a large patch size $P_L$:
\begin{equation}
\boldsymbol{T}_\text{{M}} = \text{Patchify}(\boldsymbol{z}_{\text{noised}}, P_L, c),
\end{equation}
where $\boldsymbol{T}_\text{{M}}$ are the multi-grained tokens, $c$ denotes the conditional information, \ie, class label $y$, grain map $\boldsymbol{M}$ and diffusion timestep $t$. 
These multi-grained tokens are processed by the standard DiT blocks. Following this, the routing module utilizes the grain map $\boldsymbol{M}$ to differentiate between coarse and fine tokens and restores them to latent code:
\begin{equation}
(\boldsymbol{\epsilon}_1, \boldsymbol{\epsilon}_2)= \text{Router}(\boldsymbol{T}_\text{{M}}, \boldsymbol{M}),
\end{equation}
where $\boldsymbol{\epsilon}_1,\boldsymbol{\epsilon}_2$ denote the predicted coarse noise and fine noise, respectively. 
However, a large patch size $P_L$ is not enough to handle the detailed regions. The Fine-grained Net is specifically designed to enhance noise correction in fine-grained regions, which further corrects the predicted finer-grained noise $\boldsymbol{\epsilon}_2$ using a small patch size $P_{S}$:
\begin{equation}
\boldsymbol{\epsilon}^{*}_{2} = \text{RefineNet}(\boldsymbol{\epsilon}_2, \boldsymbol{z}_{\text{noised}}, P_{S}, c),
\end{equation}
where $\boldsymbol{\epsilon}^{*}_{2}$ denotes the fine-grained noise corrected by RefineNet.
The different grained noises are then merged to create a comprehensive multi-grained noise:
\begin{equation}
\boldsymbol{\epsilon}_{\theta} = \boldsymbol{\epsilon}_{1}  \oplus \boldsymbol{\epsilon}^{*}_{2},
\end{equation}
where $\oplus$ denotes the  combination operation for the complementary noises, and $\boldsymbol{\epsilon}_{\theta}$ is the final predicted noise of D$^2$iT.
% Specifically, a smaller patchify is used to process the fine-grained region's noise, while the Noised Latent is merged as a guide and further processed by the fine-grained efficient RefineNet. Finally, the coarse and fine contents are merged to form the multi-grained noise.

\textbf{Fine-grained Efficient RefineNet.} 
The patchify operation with smaller patch size $P_{S}$ leads to more tokens and quadratic computational complexity due to self-attention.
Inspired by  \cite{liu2021swin,zhang2022styleswin,stein2024exposing}, we propose using mini-windows to divide long tokens into several short tokens.
Specifically, as illustrated in the fine patchify part of Figure \ref{fig:framework} (b), we preprocess the input fine-grained features by dividing them into multiple non-overlapping windows. Considering that global consistency is effectively modeled by the Multi-grained Net, multi-head self-attention in RefineNet is performed independently within each window, thereby effectively reducing the computational complexity. 
% This efficient refinenet can be superimposed to handle more than two granularities.

Specifically, the transformer blocks of Efficient RefineNet require effective modeling relative positional relationships of fine-grained tokens. A \textit{position-aware self-attention structure} is proposed to enhance this capability.  Specifically, a local learnable relative position bias is introduced when calculating the self-attention score in each block, enabling the model to better capture the relationship between tokens of the same granularity. The self-attention computation with the added bias is defined as:
\begin{equation}
\text{Attention}(Q, K, V) = \text{Softmax} \left( \frac{QK^\top}{\sqrt{d_k}} + B_r \right) V,
  \label{eq:attention}
\end{equation}
where \( B_r \) is an \( N \times N \) matrix representing the relative positional deviation between a given location and others. It is a set of learnable parameters updated during each training iteration, where \( N \) denotes the number of patches in the latent codes. \( Q \), \( K \), and \( V \) belong to \( \mathbb{R}^{N \times d} \) and represent the query, key, and value in the self-attention module. 

\textbf{Multi-Grained Noise Loss.}
The proposed D$^2$iT contains two granularities. To accommodate dynamic assignment, we design a multi-grained noise loss function $\mathcal{L}_{dyn}$:
\begin{equation}
  \mathcal{L}_{dyn} = \\
  \mathbb{E}_i  \mathbb{E}_{\boldsymbol{z}_0, \boldsymbol{\epsilon}_i, c, t} \left[\alpha_i \left\| \boldsymbol{\epsilon}_i - \boldsymbol{\epsilon}_{\theta_i}(\boldsymbol{z}_0, c, t \right) \right\|^2_2],
  \label{eq:dynamic_diffusion}
\end{equation}
where $\boldsymbol{z}_0$ is the latent representation of the original image.  $\alpha_i$ is the loss weight for different granularities, given by $\alpha_i = 1/(2^{k-i}\times 2^{k-i})$, where $k=2$ represents the number of granularities. $c$ denotes the conditional input that guides the diffusion process, \ie, class label and grain map. $\boldsymbol{\epsilon}_i \sim \mathcal{N} (0,I)$ represents the random noise at the $i$-th granularity. 
$\boldsymbol{\epsilon}_{\theta_i}$ is the predicted noise at the corresponding granularity. By weighting the loss according to the specified granularity and combining it with a granularity selection matrix, the multi-grained noise loss function effectively updates different noise control networks.

\section{Experiments}
\subsection{Implementation}
\begin{figure}
  \centering
   \includegraphics[width=1\linewidth]{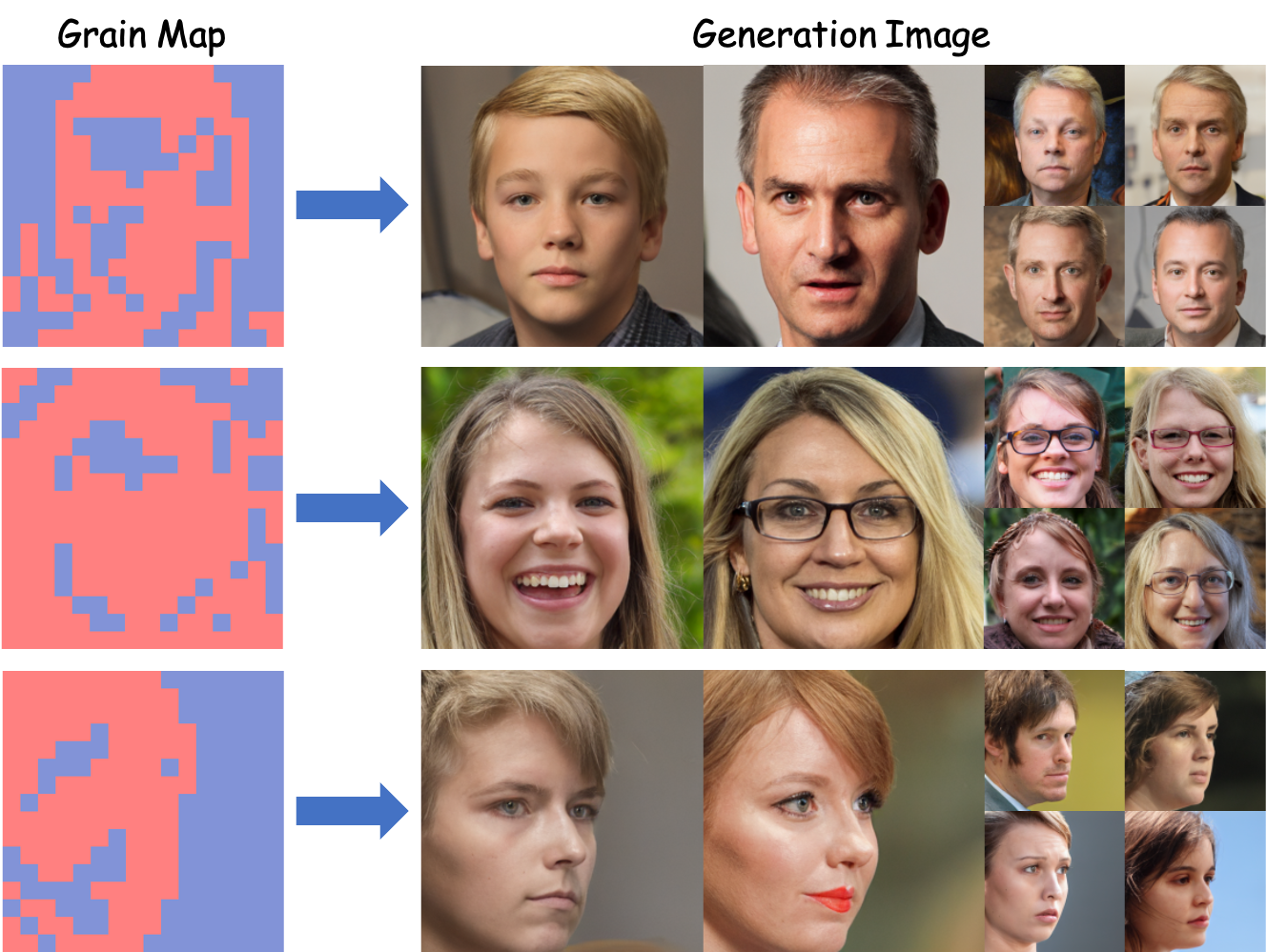}%{images/vision_res.png}
  \caption{Qualitative results of our unconditional generation on FFHQ. In the grain map, red blocks represent fine-grained regions, while blue blocks indicate coarse-grained regions.}
  \label{fig:qualitative_ffhq}
\end{figure}

\begin{table}
  \centering
  \fontsize{9}{10}\selectfont % 设置字体为 8pt，行距为 10pt
   % \resizebox{\columnwidth}{!}{%
  \begin{tabular}{@{}lccc}
    \toprule
    Model Type & Method & Param(M) & FID-10K$\downarrow$ \\
    \midrule
    GAN & VQGAN \cite{esser2021taming} & 307 & 11.4 \\
    GAN & ViT-VQGAN \cite{yu2021vector} & 738 & 13.06 \\
    VAE & VDVAE \cite{child2020very} & 115 & 28.5 \\
    Diffusion & ImageBART \cite{esser2021imagebart} & 713 & 9.57 \\
    Diffusion & UDM \cite{kim2022softtruncationuniversaltraining}  & - & 5.54 \\
    Diffusion & LDM-4 \cite{rombach2022high}  & 274 &  4.98  \\
    Diffusion & DiT-L \cite{peebles2023scalable}  & 458 &  6.26   \\ 
    \midrule
    Diffusion & \textbf{D$^2$iT-L(ours)} &  467 & \textbf{4.47} \\ 

    \bottomrule
  \end{tabular}
  \caption{Comparison of unconditional generation on FFHQ.}
  \label{tab:SOTA-FFHQ-compare}
\end{table}

\begin{figure*}
  \centering
   \includegraphics[width=1\linewidth]{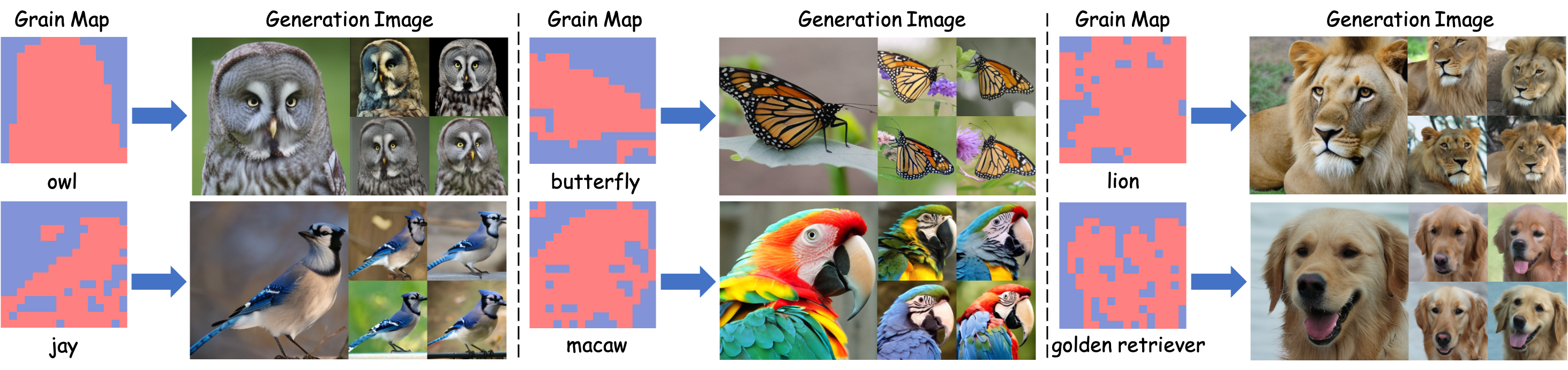}%{images/vision_res.png}
  \caption{Qualitative results of D$^2$iT-XL on ImageNet. The grain maps are generated by the Dynamic Grain Transformer based on class labels, and the images are generated by the Dynamic Content Transformer based on class labels and grain maps.}
  \label{fig:qualitative_evaluation}
\end{figure*}

We present the implementation details of DVAE and D$^2$iT, including model architecture, training details, benchmarks and evaluation metrics.

\textbf{Model architecture.} The existing SOTA DiT- based image generation models generally use $8 \times$ VAE downsampling and $2 \times$ DiT patchify operation. Therefore, in order to align with the previous work, We use two grains in pipeline. Specifically, in the first stage, the hierarchical encoder in DVAE is downsampled by factors of $F = \{8 , 16\}$ to achieve a reasonable compression, \ie, detail regions with 8$\times$ downsampling and smooth regions with 16$\times$ downsampling. Given a RGB image of shape $256 \times 256 \times 3$, it encodes a dual-grained mixture representation, \ie $32 \times 32 \times 4$ and $16 \times 16 \times 4$, where the coarse grained codes are copied and filled to the corresponding position to combine the two grains. 
In the second stage, the Dynamic Grain Transformer of D$^2$iT only needs to predict simple spatial grain distribution, so the settings remain the same as DiT-S (small model) with 33M parameters.
Dynamic Content Transformer of D$^2$iT is trained with three different settings, \ie, D$^2$iT-B (base model), D$^2$iT-L (large model), D$^2$iT-XL (extra-large model )  with 136M, 467M and 687M parameters, respectively. The Multi-grained Net and Fine-grained Net use patch sizes of 2 and 1, respectively.
The maximum time step of diffusion process is set to 1000, and a linear variance schedule of diffusion noise ranging from $10^{-4}$ to $2 \times 10^{-2}$ is used.
More detailed information of DVAE and D$^2$iT is presented in the supplementary Material.

\textbf{Training details.} 
To ensure a fair comparison with SOTA models, we adhered to the setups used in prior works:
(1) The D$^2$iT model is trained using the AdamW optimizer \cite{loshchilov2017decoupled} with a batch size of 256 and a learning rate of $1 \times 10^{-4}$.
(2) We maintain an exponential moving average (EMA) of the D$^2$iT weights with a decay rate of 0.9999 during training. 
All models are trained using eight A800 GPUs.

\textbf{Benchmarks.} Following previous work, we use two standard benchmarks, \ie,  the unconditional FFHQ \cite{karras2019style} and class-conditional ImageNet \cite{deng2009imagenet} at a resolution of 256 × 256.

\textbf{Metrics.} We use common metrics to evaluate the model. The standard Fŕechet Inception Distance (FID) \cite{heusel2017gans} is used to evaluate the generation and reconstruction quality. It measures the diversity and accuracy of the generated images. Inception Score (IS) \cite{salimans2016improved}, Precision and Recall \cite{kynkaanniemi2019improved} are also used to measure class-conditional generation on ImageNet. In order to align with previous works, we report FID-10K on FFHQ and FID-50K on ImageNet.

\subsection{Comparison Results}

We compared D$^2$iT with the state-of-the-art diffusion models on the unconditional FFHQ and class-conditional ImageNet datasets. The main results of D$^2$iT are reported using a dual-grained design of $F = \{8, 16\}$ with the ratio $r_{f=8} = 0.5$. The following results were obtained after training for 800 epochs on the respective datasets.

\textbf{Unconditional generation.} As shown in Table \ref{tab:SOTA-FFHQ-compare}, we compare our D$^2$iT-L results with existing fundamental models, achieving a $28.6\%$ improvement in quality over DiT-L along with an FID-10K score 4.47. Qualitative results for unconditional generation of D$^2$iT-L are shown in Figure \ref{fig:qualitative_ffhq}. We use a single  grain map to generate several images, and the information density distribution of the generated images is highly consistent with the grain map, demonstrating the effectiveness of our Dynamic Content Transformer.

\textbf{Class-conditional generation.}  In Table \ref{tab:imagenet_compare}, we compare the class-conditional image generation performance of D$^2$iT with existing methods. We observe that D$^2$iT outperforms DiT and other models in the table, achieving an FID score of 1.73 and a 23.8\% improvement over DiT-XL in quality with only 57.1\% of the training steps. This demonstrates the effectiveness and scalability of our model. Qualitative results of class-conditional generation are shown in Figure  \ref{fig:qualitative_evaluation}.  %More visual results are available in the supplementary materials.

% Table \ref{tab:SOTA-compare} compares our DG-DiT with standard DiT at different model sizes. DG-DiT achieves higher FID scores at both model scales with less training cost. The parameters of DG-DiT are similar to those of DiT.

\begin{table}
  \centering
    \setlength{\tabcolsep}{2pt} % 设置列间距为2pt
    \renewcommand{\arraystretch}{0.95}
   \resizebox{\columnwidth}{!}{%
  \begin{tabular}{@{}lcccccc}
    \toprule
    Method & Param(M) & FID-50K$\downarrow$ & IS$\uparrow$ & Prec.$\uparrow$ & Rec.$\uparrow$ \\
    \midrule
    % VQVAE-2 \cite{razavi2019generating} &   & 31.11 & - & 0.36 & 0.57   \\
    VQGAN \cite{esser2021taming} & 397  & 15.78 & 78.3 & - & -   \\
    % StyleGAN \cite{karras2020analyzing} &     & 2.30 & 265.1 & 0.78 & 0.53 \\
    MaskGIT \cite{chang2022maskgit}  &  227  & 7.32 & 156.0 & 0.78 & 0.50 \\
    DQ-Transformer \cite{huang2023towards} &  655  & 5.11 & 178.2 & - & - \\
    LlamaGen \cite{sun2024autoregressive}  & 775 & 2.62  &  244.1 &  0.80 & 0.57  \\
    DiGIT \cite{zhu2024stabilize} & 732 & 3.39 & 205.96 & - & - \\
    Open-MAGVIT2-XL \cite{luo2024open}& 1500 & 2.33 & 271.77 & 0.84 & 0.54 \\
    VAR \cite{tian2024visual}  & 600 & 2.57 &  302.6  & 0.83 & 0.56  & \\
    % TiTok-B-64 \cite{yu2024image}  &&&&&& \\
    \midrule
    ADM \cite{dhariwal2021diffusion} &  554  & 10.94 & 100.98 & 0.69 & 0.63 \\
    LDM-4 \cite{rombach2022high} &  400  & 10.56 & 103.49 & 0.71 & 0.62  \\
    DiT-XL \cite{peebles2023scalable} &  675  & 9.62 & 121.50 & 0.67 & 0.67  \\
    % RDM \cite{teng2023relay} &848 & 5.27 & 153.43 & 0.75 & 0.62 & \\
   MDT\cite{Gao_2023_ICCV} & 676 & 6.23 & 143.02 & 0.71 & 0.66 \\
    \textbf{D$^2$iT-XL(ours)} & 687 & 5.74 & 156.29  & 0.72  & 0.66 \\ %6.73
    \midrule
    ADM-G \cite{dhariwal2021diffusion} &  554  & 4.59 & 186.70 & 0.82 & 0.52 \\
    ADM-G-U \cite{dhariwal2021diffusion} &  554  & 3.94 & 215.84 & 0.83 & 0.53 \\
    LDM-4-G \cite{rombach2022high}  &  400 & 3.60 & 247.67 & 0.87 & 0.48  \\
    % U-ViT-G \cite{bao2022all}  &   & 3.40 & - & - & -  \\
    DiT-XL-G \cite{peebles2023scalable}  & 675  & 2.27 & 278.24 & 0.83 & 0.57  \\
    RDM-G \cite{teng2023relay} & 848 & 1.99 & 260.45 & 0.81 & 0.58 & \\
    DiMR \cite{liu2024alleviating} & 505 & 1.70 & 289.0 &  0.79 & 0.63 & \\
    % f-DM \cite{gu2022f} &&&&&& \\
    MDT-G\cite{Gao_2023_ICCV} & 676 & 1.79 & 283.01 & 0.81 & 0.61 \\
    MDTv2-G\cite{Gao_2023_ICCV} & 676 &  1.58 & 314.73 & 0.79 &  0.65 \\
    \textbf{D$^2$iT-XL-G(ours)} & 687 & 1.73 & 307.89  & 0.87  & 0.56 \\
    \bottomrule
  \end{tabular}}
  \caption{Comparison of class-conditional generation on ImageNet $256 \times 256$. -G indicates the results with classifier-free guidance.}
  \label{tab:imagenet_compare}
\end{table}

\subsection{Ablation Study and Analysis}
In this section, we conduct ablation studies to validate the design of D$^2$iT. We report the results of the D$^2$iT-B model on the FFHQ dataset, using FID-10K as the evaluation metric unless otherwise stated. All D$^2$iT-B models in the ablation study were trained for 50 epochs.

\begin{figure}[t]
  \centering
   \includegraphics[width=1\linewidth]{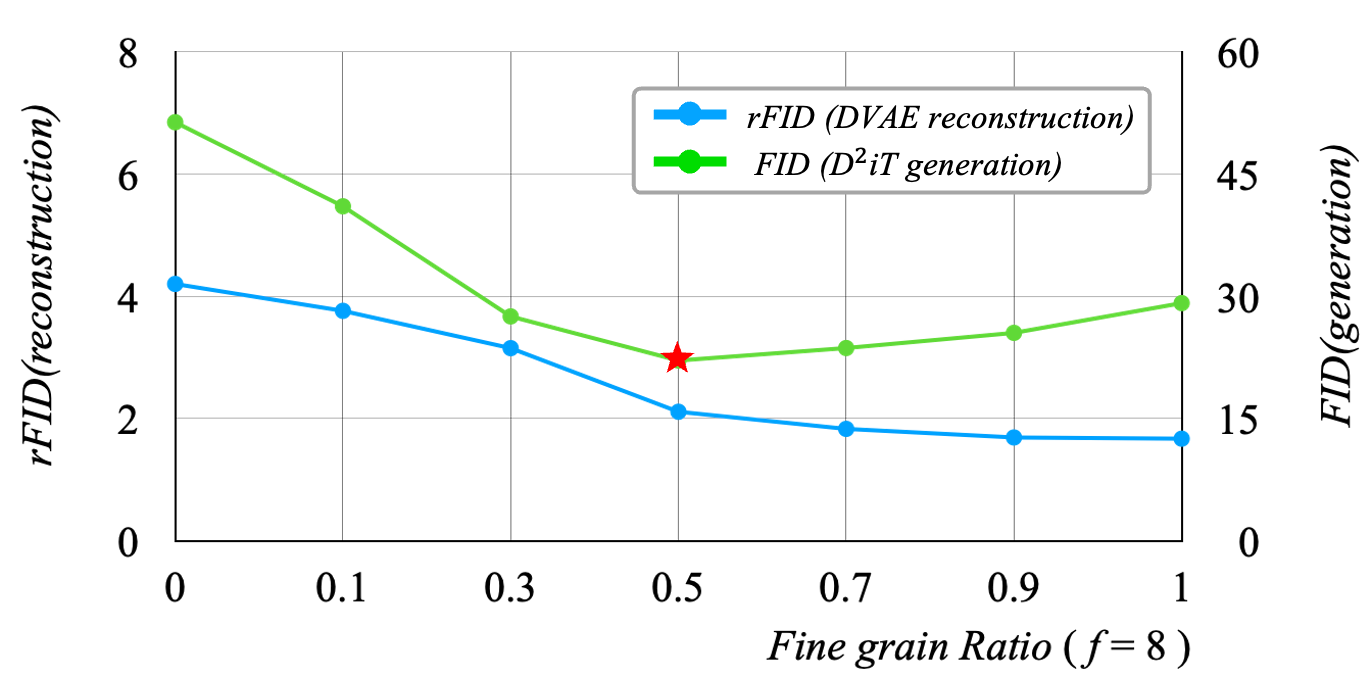}
   \caption{The curves of different grain ratios of reconstruction  quality (rFID) to generation quality (FID) on FFHQ.}
   \label{fig:ffhq-ratio}
\end{figure}

\textbf{Dynamic Granularity Strategy.} We first demonstrate that our dynamic granularity noise diffusion has better generation ability compared to the existing fixed noise diffusion. We use VAE with $f=16$ and $f=8$ as baselines, and DVAE uses $F=\{8,16\}$ dual granularity. Figure \ref{fig:ffhq-ratio} shows the line graph of the reconstruction ability of DVAE and the generation ability of D$^2$iT when the proportions of different granularities vary.  We could conclude that: (1) As the proportion of fine granularity increases, the reconstruction quality of DVAE gradually improves because more codes can better represent the image. (2) At the appropriate fine grain ratio, D$^2$iT shows better image generation ability (FID is 22.11 when $r_{f=8}=0.5$) compared to the fixed level noise(FID is 29.15 when $r_{f=8}=1$ and FID is 51.33 when $r_{f=8}=0$). The reason is that important regions require more coding, \ie, more noise representation, while less important regions suffice with less noise due to their lesser information. (3) When the fine grain ratio increases from 0.7 to 1.0, DVAE only gets 0.16 improvement (from 1.83 to 1.67) in rFID, but the performance of D$^2$iT declines from 23.64 to 29.15, indicating that the last 30\% of less important regions contribute little effective information to the image, and most of it is redundant. Consequently, using too much code to represent coarse-grained regions hinders the model's performance. The experimental results strongly support the motivation of dynamic diffusion to eliminate both insufficiency and redundancy.

\begin{figure}
  \centering
   \includegraphics[width=1\linewidth]{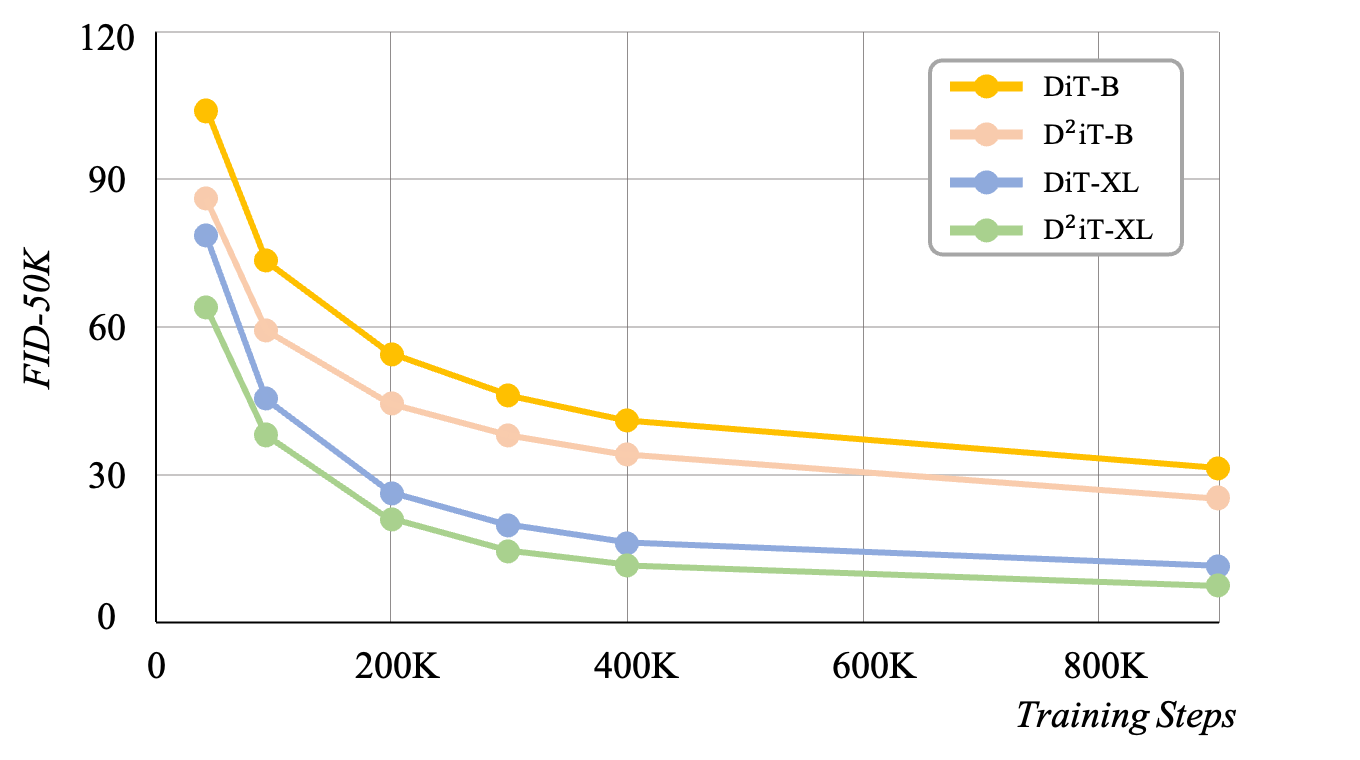}%{images/vision_res.png}
  \caption{Training convergence comparison of DiT and our D$^2$iT with different parameters on ImageNet. FID-50K is evaluated.}
  \label{fig:Training_convergence}
\end{figure}

In addition, Figure \ref{fig:Training_convergence} shows the training convergence between our method and DiT with different parameters on the ImageNet dataset. D$^2$iT shows faster convergence than DiT in models with similar parameters.

\textbf{Analysis on the enhanced design.}
% We first verify the design of different detail components in Table 7. The position prediction network provides important granularity information for the second stage generation, helping the model to distinguish different granularities. Without this module, DG-DiT cannot know which granularity code should be generated next, so the performance will be worse. Then we distinguished the losses of different granularities. The coarse-grained areas are represented by repeated codes, so corrections need to be made when calculating the loss to achieve the best performance. In addition, the introduction of relative position encoding allows the multi-granularity processing module to distinguish the relative positions of coarse and fine-grained areas, which helps modeling at different positions. The fine-grained processing network uses a smaller patch size to further process the detail area and enrich the representation of the details, which also improves the image generation capability.
We first verify the design of various detailed components in Table \ref{ablation_configuration}. Initially, DVAE and the Routing Module are introduced based on the standard VAE \& DiT architecture to implement simple dynamic grain diffusion. It was observed that the FID score improved from 34.67 to 29.10. Subsequently, by distinguishing different losses in coarse-grained and fine-grained regions, we further improved the FID score to 27.62. Next, we replaced the last two standard DiT backbone layers with RefineNet layers that have smaller patch size and learnable position embeddings, ultimately achieving an FID score of 22.11, compared to DiT’s 34.67.

\textbf{Analysis on the effectiveness of the RefineNet.}
Table \ref{Swin-T_ablation} presents two sets of control experiments to verify the effectiveness of Efficient RefineNet. (1) We adjusted the number of DiT Blocks and RefineNet Blocks, keeping the model size and Gflops\footnote{Gflops (Giga Floating Point Operations per Second) is a metric to describe the computational complexity of a model.} relatively constant, and discovered that processing the detail regions can further enhance generation capability. This illustrates that the RefineNet with small patch size can capture richer details. In addition, we cannot blindly increase the number of RefineNet layers under certain computing resources. Too few layers in DiT backbone will damage global consistency and the generation of coarse-grained regions.  Setting the appropriate ratio of layers, \ie five DiT layers to one RefineNet layer, proves beneficial for generation under fixed computing resources. (2) We examined the impact of increasing the number of RefineNet blocks while maintaining a constant number of DiT blocks, demonstrating RefineNet’s scaling capabilities.

% \begin{table}
%   \centering
%   \begin{tabular}{@{}lcc}
%     \toprule
%     Method & Training Steps(k) & FID-50K$\downarrow$ \\
%     \midrule
%     DiT-B/2  & 400 & 43.47 \\
%     D$^2$iT-B(ours)  & 400 & \textbf{37.16} \\
%     D$^2$iT-B(ours)  & 1000 & \textbf{22.41} \\
%     \midrule
%     DiT-XL/2 &  400 & 19.47 \\
%     DiT-XL/2 &  2352 & 10.67 \\
%     DiT-XL/2 &  7000 & 9.62 \\
%     D$^2$iT-XL(ours)  & 400 & \textbf{16.68} \\
%     D$^2$iT-XL(ours)  & 2500 & \textbf{8.82} \\
%     \bottomrule
%   \end{tabular}
%   \caption{Comparison of the performance of Base (B) and Extra Large (XL) size models on the ImageNet dataset at different training steps.}
%   \label{tab:SOTA-compare}
% \end{table}

\begin{table}[t]
\centering
\fontsize{8.8}{8}\selectfont % 设置字体为 8pt，行距为 10pt
\begin{tabular}{lc}
\toprule
Configuration  & FID-10K$\downarrow$ \\
\midrule
DiT-B & 34.67\\
\midrule
with Predicted Grain Map  \\
+ DVAE  \& Routing Module  & 29.10 \\
+ Dynamic grain loss  & 27.62 \\
% + Learnable position embd  & 25.15 \\
+ RefineNet \& Learnable Pos-embed (ours D$^2$iT)  & 22.11 \\
\bottomrule
\end{tabular}
\caption{Ablation study of D$^2$iT-B on FFHQ. All additional designs use  grain map predicted by Dynamic Grain Transformer.}
\label{ablation_configuration}
\end{table}
% \begin{table}[ht]
% \centering
% \begin{tabular}{lll}
% \toprule
% Configuration & Setting & rFID$\downarrow$ \\
% \midrule
% DiT-B/2 & & 34.67\\
% \midrule
% + DG-VAE & $\times$ & -\\
%                     & $\checkmark$ & 29.10 \\
% \midrule
% + Dynamic grain loss  & $\times$ & 29.10 \\
%                     & $\checkmark$ & 27.62 \\
% \midrule
% + Learnable position embd  & $\times$ & 27.62 \\
%                     & $\checkmark$ & 25.15 \\
% \midrule
% + Fine processing network   & $\times$ & 25.15 \\
%                     & $\checkmark$ & 22.11 \\

% \bottomrule

% \end{tabular}
% \caption{Ablation study on DG-DiT-B in FFHQ. Models are trained for 50 epochs.}
% \end{table}

\begin{table}[ht]
\centering
\fontsize{9}{8}\selectfont % 设置字体为 8pt，行距为 10pt
\begin{tabular}{cccc}
\toprule
Total Layers&DiT & RefineNet & FID-10K$\downarrow$ \\
\midrule
% \multicolumn{3}{l}{Fixed number of Total Blocks.} \\ % 在这里添加文本
12& 12 & 0 & 25.15\\
12& 10 & 2 & \textbf{22.11}\\
12& 8 & 4 & 26.53\\
\midrule
% \multicolumn{3}{l}{Fixed number of DiT Blocks.} \\ % 在这里添加文本
12& 12 & 0 & 25.15\\
14& 12 & 2 & 20.99\\
16& 12 & 4 & 19.96\\
\bottomrule
\end{tabular}
\caption{Effect of numbers of RefineNet Blocks in D$^2$iT-B. Experiments with fixed total layers increasing Refinenet layers and fixed DiT layers increasing Refinenet layers.}
\label{Swin-T_ablation}
\end{table}
\begin{table}[ht]
\centering
\fontsize{9}{8}\selectfont % 设置字体为 8pt，行距为 10pt
\begin{tabular}{lc}
\toprule
 Grain Map Setting & FID-10K$\downarrow$ \\
\midrule
 Random & 15.93 \\
 Ground Truth & 4.35 \\
 Dynamic Grain Transformer & 4.47 \\
\bottomrule
\end{tabular}
\caption{Effect of Dynamic Grain Transformer with D$^2$iT-L.}
\label{DGT_ablation}
\end{table}

\textbf{Analysis on Dynamic Grain Transformer.}
We present the results of D$^2$iT-L experiments with various grain map settings after 800 training epochs on the FFHQ dataset in Table \ref{DGT_ablation}. The grain map generated by the Dynamic Grain Transformer yields results comparable to the Ground Truth grain map of the datasets, significantly outperforming the random grain map. This demonstrates that the lightweight Dynamic Grain Transformer is sufficient to accurately model the spatial distribution of real images. 
\section{Conclusion \& Future Direction}
In this study, we point out that existing Diffusion Transformer (DiT) models apply fixed denoising to uniformly sized image regions, disregarding the naturally varying information densities across different regions. This oversight results in insufficient denoising in crucial regions and redundant processing in less significant ones, compromising both local detail authenticity and global structural consistency. To address these limitations, a novel two-stage framework is designed to enhance the effectiveness of image generation.
% (1) DVAE which uses a hierarchical encoder to encode images at different granularities tailored to their specific information densities. (2) D$^2$iT which generates images by predicting different levels of noise for coarse-grained (less noise for smooth regions) and fine-grained (more noise for detailed regions), through a novel dynamic granularity architecture. In addition, the Efficient RefineNet achieves better balance between computational requirements and visual fidelity. Comprehensive experiments on various generation tasks validate the effectiveness of our approach.
% D$^2$iT generates images by predicting multi-grained noise through a Dynamic Grain Transformer and Dynamic Content Transformer pipeline. 
The effectiveness of our method is demonstrated in various comprehensive generative experiments.

\textbf{Future Direction.}  Our study confirms the effectiveness of the dynamic granularity strategy in diffusion process and uses two granularities to align with existing methods. In future work, more granularities within the dynamic diffusion transformer architecture can be explored.
{
    \small
    \bibliographystyle{ieeenat_fullname}
    \bibliography{main}
}
% \maketitle
% \input{paper_section/X_suppl}
% % WARNING: do not forget to delete the supplementary pages from your submission 
% {
%     \small
%     \bibliographystyle{ieeenat_fullname}
%     \bibliography{main}
% }

\end{document}